\documentclass[letterpaper, 10 pt, conference]{ieeeconf}  % Comment this line out if you need a4paper
\IEEEoverridecommandlockouts                              
\usepackage{amsmath, amssymb, graphicx, hyperref}
\usepackage{xcolor}
\usepackage{tikz}
\usetikzlibrary{shapes, arrows.meta, positioning, fit}

\newcommand\copyrighttext{%
\footnotesize \copyright 2024 IEEE. Personal use of this material is permitted. Permission from IEEE must be obtained for all other uses, in any current or future media, including reprinting/republishing this material for advertising or promotional purposes, creating new collective works, for resale or
redistribution to servers or lists, or reuse of any copyrighted component of this work in other works.}
\newcommand\copyrightnotice{%
\begin{tikzpicture}[remember picture,overlay]
\node[anchor=north,yshift=-20pt] at (current page.north) {\fbox{\parbox{\dimexpr\textwidth-\fboxsep-\fboxrule\relax}{\copyrighttext}}};
\end{tikzpicture}%
}

\title{\LARGE \bf
Guess the Drift with LOP-UKF: LiDAR Odometry and Pacejka Model for Real-Time Racecar Sideslip Estimation    
}

\author{Alessandro Toschi$^{1*}$, Nicola Musiu$^{1*}$, Francesco Gatti$^{2}$, Ayoub Raji$^{1}$, \\
Francesco Amerotti$^{2}$, Micaela Verucchi$^{2}$ and Marko Bertogna$^{1}$% <-this % stops a space
\thanks{$^{1}$University of Modena and Reggio Emilia, Italy,\newline
        {\tt\small \{alessandro.toschi, nicola.musiu, ayoub.raji, marko.bertogna\}@unimore.it}}
\thanks{$^{2}$Hipert srl, Italy,\newline
        {\tt\small \{francesco.gatti, francesco.amerotti, micaela.verucchi\}@hipert.it}}%
\thanks{$*$The authors contributed equally.}
}

\begin{document}
\maketitle
\thispagestyle{empty}
\pagestyle{empty}

\copyrightnotice
%%%%%%%%%%%%%%%%%%%%%%%%%%%%%%%%%%%%%%%%%%%%%%%%%%%%%%%%%%%%%%%%%%%%%%%%%%%%%%%%
\begin{abstract}
The sideslip angle, crucial for vehicle safety and stability, is determined using both longitudinal and lateral velocities. However, measuring the lateral component often necessitates costly sensors, leading to its common estimation, a topic thoroughly explored in existing literature. This paper introduces LOP-UKF, a novel method for estimating vehicle lateral velocity by integrating Lidar Odometry with the Pacejka tire model predictions, resulting in a robust estimation via an Unscendent Kalman Filter (UKF). This combination represents a distinct alternative to more traditional methodologies, resulting in a reliable solution also in edge cases. We present experimental results obtained using the Dallara AV-21 across diverse circuits and track conditions, demonstrating the effectiveness of our method. 
\end{abstract}

\section{INTRODUCTION}
\label{section:intro}

The lateral dynamics of vehicles play a vital role in understanding and predicting vehicular behavior across different driving conditions. Accurate estimation of lateral velocity is crucial for calculating the sideslip angle, defined as the angle between the vehicle's orientation and its actual direction of motion. Precise knowledge of this angle is the key to ensuring vehicle stability and control, especially in challenging driving scenarios.\\
The high costs associated with direct measurement sensors necessitate the estimation of lateral velocity through advanced algorithms, using more affordable, readily available sensors.
Section \ref{section:application} will delve deeper into the reasons why Advanced Driver Assistance Systems (ADAS) and Autonomous Vehicles (AV) benefit from this estimation.\\
\begin{figure}
    \centering
    \includegraphics[width=1.0\linewidth]{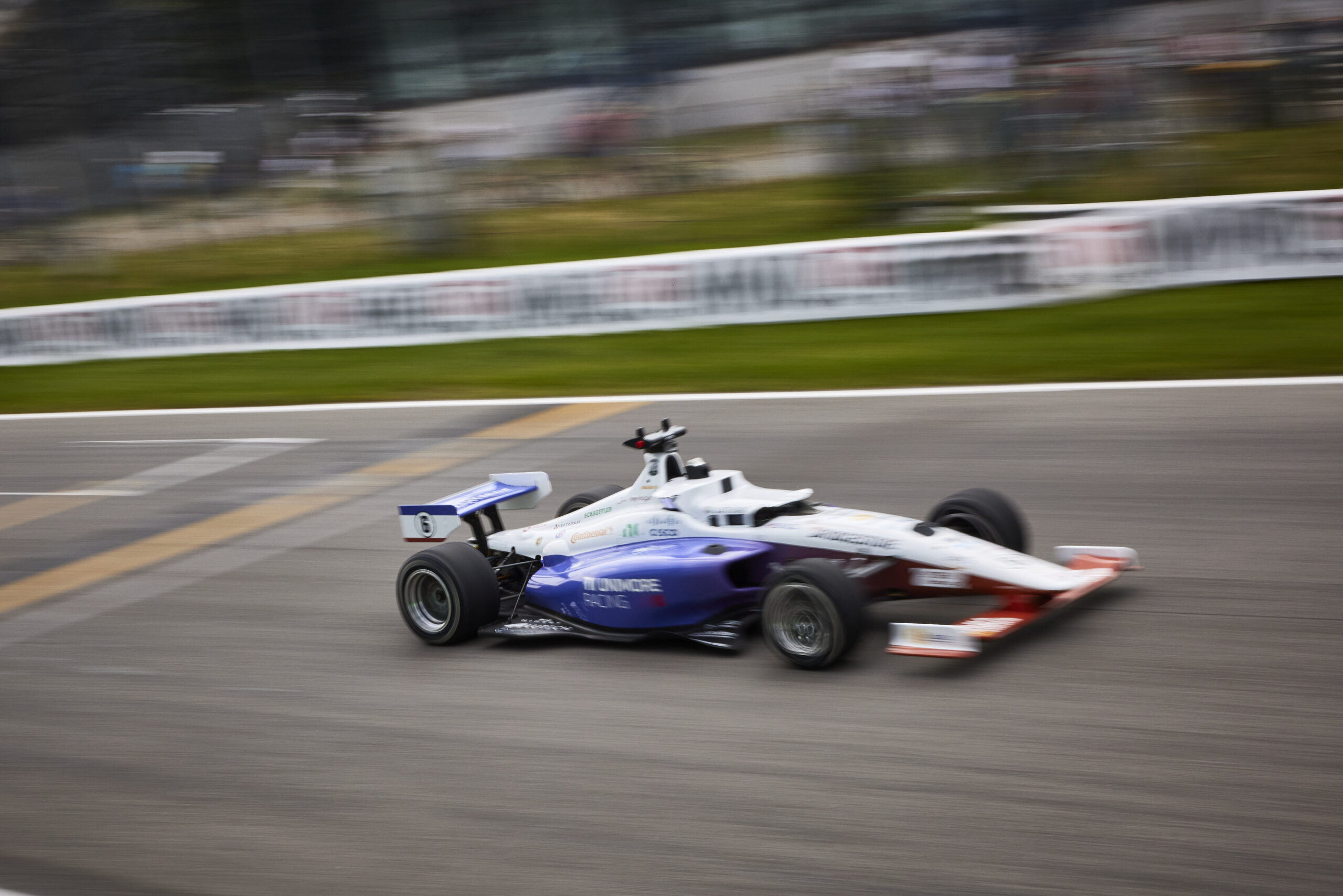}
    Dallara AV-21 - TII UNIMORE Racing during the Indy Autonomous Challenge at MIMO 2023 at the Monza F1 circuit.
    \label{fig:indy_monza}
\end{figure}
Our approach, the LOP-UKF filter, is designed to estimate three key parameters pertaining the lateral dynamics of the car: lateral velocity, lateral acceleration, and yaw rate. To achieve this, it utilizes a combination of sensors, either directly or indirectly. These include two Inertial Measurement Units (IMUs), two Global Navigation Satellite Systems (GNSSs), a steering encoder, phonic wheels, and three LiDARs. In parallel, the Kistler SF-Motion optical sensor, employing non-contact technology, measures directly the lateral velocity of the vehicle. The data of this sensor, while not used in the filter processing, is essential for post-analysis comparison, serving as a benchmark to verify the accuracy of the filter estimations. The vehicle equipped with these sensors is the Dallara AV-21 Fig.\ref{fig:indy_monza} used in the Indy Autonomous Challenge (IAC{\footnote{\href{https://www.indyautonomouschallenge.com/}{https://www.indyautonomouschallenge.com/}}), an international autonomous racing competition among universities in which we participate as TII Unimore Racing team. The results obtained on different tracks, with different vehicle setups and grip conditions will be presented in Section \ref{section:results}\\

\section{RELATED WORK}
\label{section:related}
Plenty of research has been devoted to estimating the vehicle sideslip angle across different contexts and scales. These methods broadly fall into two categories: Neural Networks (NN) and model-based observers \cite{chindamo2018vehiclesideslip}.
Kinematic approaches, such as the one proposed in \cite{selmanaj2017vehiclesideslip}, can be a simple solution for estimating the sideslip. Later, hybrid methodologies combining kinematic and dynamic models \cite{carnier2023hybrid, villano2021crosscombined} have demonstrated enhanced accuracy and adaptability.
The Extended Kalman Filter (EKF) and the Unscented Kalman Filter (UKF) are two principal methods for state estimation in nonlinear systems, each with distinct advantages depending on the application context. The EKF, primarily used in systems with mild nonlinearity, linearizes the state transition and observation models through Jacobian matrix calculations. This approach, while computationally efficient, may lead to inaccuracies in highly nonlinear scenarios \cite{wan2000unscented, heidfeld2019ukfbased}. However, in linear regions or situations with small nonlinearities, the EKF's performance is comparable to that of the UKF \cite{dinverno2021benchmark}. On the other hand, the UKF avoids the need for linearization as will be explained in Section \ref{section:ukf}. Despite its increased computational demand, the UKF's ability to more accurately handle severe nonlinearities often justifies its use, especially in complex dynamic systems where precision is paramount \cite{alshawi2024adaptive}.
Regarding the other macro category for sideslip estimation, NNs have occasionally outperformed traditional state-of-the-art solutions in racing environments \cite{srinivasan2020endtoend}, proving effective even under varying tire wear and grip conditions in road vehicles \cite{giuliacci2023recurrent}. Neural Networks have the potential to provide more accurate estimations while requiring less computational resources. However, they require extensive training and may struggle to adapt to new system conditions \cite{chindamo2021experimental}. In essence, NNs for state estimation work as observers based on highly precise empirical models. The drawback is that the resulting black box is not parametrizable, thus limiting its generalizability and adaptability to different vehicles and environments. On the other hand, the EKF and UKF offer reliable estimations but become computationally intensive as the model precision increases. Additionally, Kalman filters inherently incorporate the concept of uncertainty, which proves advantageous when employed within more complex systems, as they do not only provide a quantity, but also indicate the confidence level of the estimation.
Finally, it is worth citing a few approaches that do not fall into the two categories above and highlight the emerging role of computer vision in vehicle dynamics. The work in \cite{harchut2009camera} presents a cost-effective technique using standard vehicle sensors and an automotive camera. In \cite{SerenaVision} the phase correlation algorithm is explored, and \cite{kuyt2018mixed} developed a method combining kinematic data with camera measurements, both demonstrating computer vision potential in sideslip angle estimation in RC cars
The proposed LOP-UKF, particularly tailored for racing environments, uses a UKF to integrate the Pacejka tire model sec.\ref{section:pacejka} with LiDAR Odometry sec.\ref{section:LiDAR}. The latter makes the estimation more robust when the conditions differ from those fitted by the tire model. To the best of our knowledge, this is a novel contribution to the actual literature on the estimation of the sideslip angle and the lateral velocity.

\section{APPLICATION}
\label{section:application}

As emphasized in \cite{napolitano2024fourwheeled}, vehicle status estimations impact a wide range of systems and are vital for both AVs and human-driven vehicles. The main objective in ADAS and autonomous driving is to enhance the safety and comfort of individuals inside the vehicle. On the other hand, motorsports focus on maximizing the performance of the vehicle. In the dynamic field of autonomous racing, the role of accurate lateral velocity estimation becomes even more critical. It facilitates controlled yet aggressive maneuvering, particularly in cornering scenarios, allowing vehicles to maintain high speeds without compromising safety and stability. In this context the objectives of maximizing vehicle performance and ensuring safety intersect, with a focus on the car's integrity in the absence of a human driver.
\begin{figure}
	\centering
	\includegraphics[clip, trim=4cm 9cm 4cm 9cm, width=1.0\columnwidth]{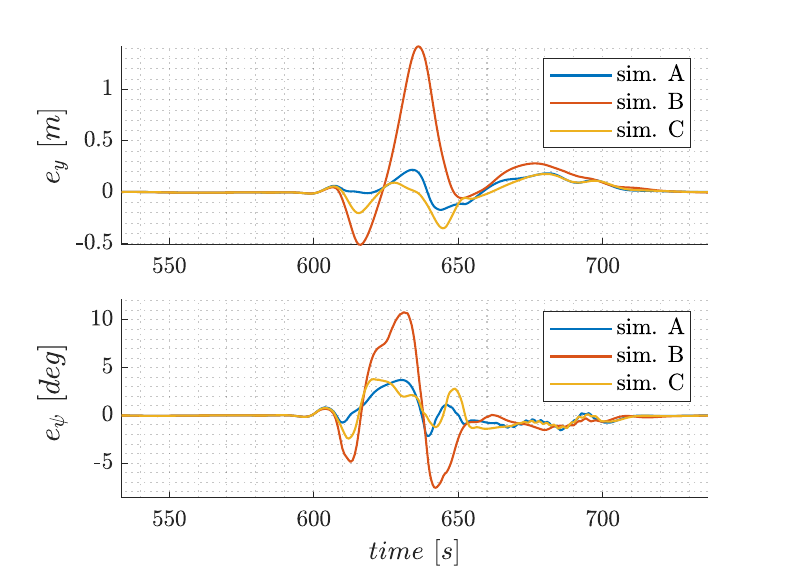}
    \caption{Lateral and heading tracking error of the Model Predictive Controller in the simulator. Here, Prima Variante turn of the Monza circuit is shown.}
	\label{fig:mpc_vy}
\end{figure}
The algorithm proposed in this work arises mainly to support the Model Predictive Controller (MPC), developed in \cite{raji2023tricycle}, in estimating its state, in which lateral speed and yaw rate also appear. A simulation environment based on a multi-body model with highly accurate vehicle dynamics was exploited to test the goodness of the solution and the interaction between the modules. Integrating the lateral velocity feedback into the controller improves the path tracking accuracy, as shown in Fig. \ref{fig:mpc_vy}:
In simulation A (blue line), the state vector of the controller is updated with $v_y$, which represents the ground truth provided by the simulator. In simulation B (orange line), no feedback is forwarded to the controller, and hence $v_y$ is initialized to 0. In simulation C (ochre line), the controller is fed with a $v_y$ which is the $70$\% of the actual value.
This experiment demonstrates the importance of having an accurate estimate of the lateral velocity, both in terms of safety and performance: for instance, in the first simulation, the lateral error is within the range of $20 cm$, whereas when the state is not initialized it is close to $1.5 m$. For this application, this amount of error can translate to off-track or hitting any obstacle, leading the machine into a potentially unsafe condition.
Importantly, the results of the third simulation demonstrate that capturing the overall trend, even with a significant absolute error, can help the controller maintain the tracking error within an acceptable limit, even under extreme conditions. Indeed, a lateral acceleration of around $15 m/s^2$ has been reached.

\section{UNSCENTED KALMAN FILTER}
\label{section:ukf}

As discussed in Section \ref{section:related}, the UKF is a viable solution to estimate the lateral dynamics of a vehicle because it does not linearize the state function at the working point but approximates its probability distribution. This is done using the unscented transformation, a deterministic sampling technique that employs sigma points. 
In this section, we provide a comprehensive explanation of the two stages of the process: prediction and update \cite{wan2000unscented}. The aim is to clarify the utilization of the Pacejka model for prediction and the application of LiDAR odometry for the update.
The model is described by the state vector $X$, which contains the lateral velocity $v_y$, the yaw rate $r$ and the lateral acceleration $a_y$:\\
\begin{equation} \label{eq:state}
    X = \begin{bmatrix}
    v_y; 
    r; 
    a_y \end{bmatrix}
\end{equation}
The input variables are the steering angle $\delta$, the longitudinal velocity $v_x$, the longitudinal acceleration $a_x$ and the banking angle $\theta$:
\begin{equation} \label{eq:input}
    u = \begin{bmatrix}
    \delta, 
    v_x, 
    a_x, 
    \theta \end{bmatrix}
\end{equation}

\subsection{Prediction Phase}

\subsubsection{Sigma Points Generation}
Given a state estimate $\hat{X}$ and covariance $P$, a set $\mathcal{X}$ of sigma points are generated to capture the mean and covariance of the state distribution. The UKF employs various calibration parameters to define the dispersion of sigma points and to describe the probability distribution tail heaviness and shape. The vector $\mathcal{X}$ has the same dimension as the state vector, for which a sigma-point is a potential guess. The notation $\mathcal{X}_{k}^{(i)}$ denotes the i-th sigma-point at the time step k. The number of the sigma points is linearly dependent on the state dimension $N$.
\subsubsection{Sigma Points Transformation}
The sigma points and input variables are passed through the nonlinear state transition function $f$, resulting in transformed points $Y^{(i)}$, capturing the non-linearity of the process.
\begin{align} \label{model_update}
    Y^{(i)} = \mathcal{X}_{k+1|k}^{(i)} &= f(\mathcal{X}_{k}^{(i)}, u_k), \quad \text{for } i = 0, \ldots, 2N.
\end{align}
Details about the implemented $f$ function will be given in the next section.
\subsubsection{Predicted State and Covariance Calculation}
From the transformed sigma points, the predicted state $\hat{X}'$ and covariance $\hat{P}'$ are calculated, representing the posterior distribution after the non-linear transformation.
\begin{align*}
    &\hat{X}' = \hat{X}_{k+1|k} = \sum_{i=0}^{2N} W_m^{(i)} Y^{(i)}, \\
    &P_{k+1|k} = \sum_{i=0}^{2N} W_c^{(i)} (Y^{(i)} - \hat{X}')(Y^{(i)} - \hat{X}')^T + Q,
    %&\text{where } W_m^{(i)} \text{ and } W_c^{(i)} \text{ are weights for the mean and covariance }
\end{align*}
where $W_m^{(i)}$ and $W_c^{(i)}$ are weights for mean and covariance, dependent on the same UKF calibrations parameters discussed above. The diagonal matrix $Q$ contains the variances of the process noise, essentially representing parameters that determine the level of confidence we place in the model.
\begin{figure}
    \centering
    \includegraphics[width=1\linewidth]{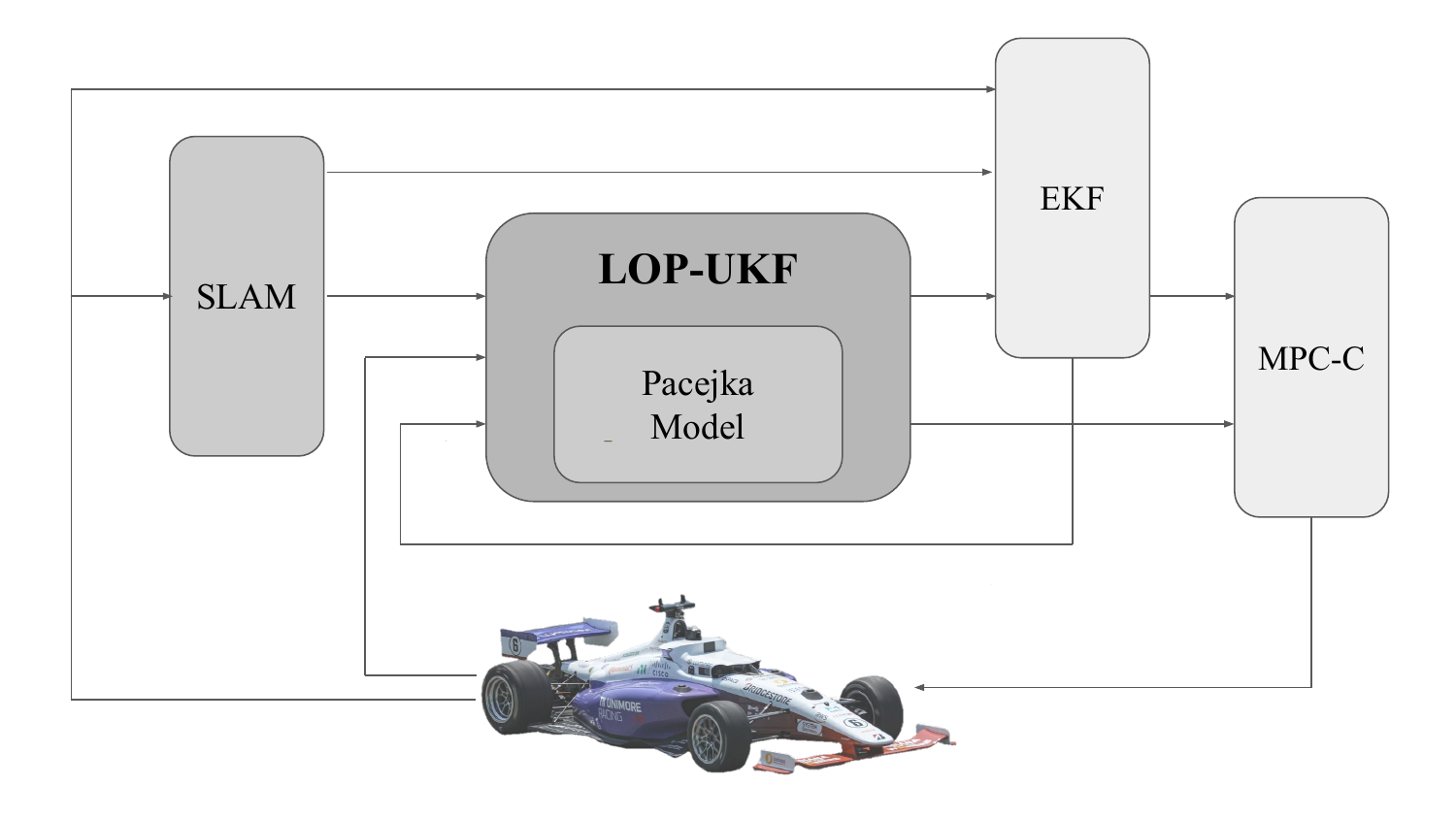}
    \caption{Diagram showing LOP-UKF interaction with other key parts of the autonomous driving system}
    \label{fig:ukf_scheme}
\end{figure}

\subsection{Update Phase}
\subsubsection{Measurement Incorporation}
The sigma points are then propagated through the measurement function $h$:
\begin{align*}
    &Z_{k}^{(i)} = h(Y^{(i)}), \quad \text{for } i = 0, \ldots, 2L.\\
    &\hat{Z}_k' = \hat{Z}_{k+1|k} = \sum_{i=0}^{2N} W_m^{(i)} Z_{k}^{(i)}
\end{align*}
where $\hat{\mathbf{Z}}'$ is the predicted measurement. In our implementation, two distinct functions $h$ are used. One is implemented for the LiDAR input and one for the two IMUs. These are the three sources used in the filter state: $v_y$ is received just from the LiDAR, $a_y$ from the two IMUs, and $r$ from all the sensors.
Once $\hat{Z}'$ is calculated, the covariances are updated: $\hat{\mathbf{P}}'_{xz}$ is the cross-covariance between the state and the measurement, and $\hat{\mathbf{P}}'_{zz}$ is the predicted measurement covariance.
The matrix $\hat{\mathbf{P}}'_{zz}$ is where the variances of the measurement noise are contained through the diagonal matrix $R$. Upon receiving a new measurement $\mathbf{Z}$, the filter updates the state estimate and covariance. This involves calculating the Kalman gain $K$ and updating both the state estimate $\hat{X}$ and the covariance estimate $P$:
\begin{align*}
    &K_k = P_{x_kz_k} P_{z_kz_k}^{-1}, \\
    &\hat{X}_{k+1} = \hat{X}' + K_k(Z - \hat{Z}_{k}'), \\
    &P_{k+1} = P_k - K_k P_{z_kz_k} K_k^T.
\end{align*}

\subsection{Implementation}
\label{lop-ukf-implementation}

A schematic of the interaction between our autonomous driving software and the vehicle, specifically focusing on the LOP-UKF implementation, is shown in Fig. \ref{fig:ukf_scheme}.
An updated version of the EKF discussed in \cite{er-autopilot}, is used to estimate the vehicle's position, orientation, longitudinal velocity, and acceleration. The LOP-UKF and the EKF, exchanging lateral and longitudinal velocities, were both developed within Simulink. Their integration into our autonomous driving stack was facilitated through a ROS2 wrapper after converting Simulink models into C++ code. The resulting solution has a low computational load and works smoothly at 125 Hz.
Additionally, our system incorporates a Simultaneous Localization And Mapping (SLAM) module, as detailed in Section \ref{section:LiDAR}. The SLAM outputs vehicle status estimations, which are utilized by the EKF and LOP-UKF.
Finally, the output of these two filters fills the state of the MPC. It's important to note that despite the potential variance in the update rates of the Kalman filters' data sources, such differences do not adversely impact the functionality of the filters themselves.

\section{PACEJKA TIRE MODEL}
\label{section:pacejka}

To achieve accurate estimation while maintaining low complexity, a single-track model has been employed to represent the lateral dynamics.
The lateral forces that the tires can exploit are modeled using the semi-empirical Magic Formula \cite{pacejka}, wherein first guess parameters are derived through offline fitting based on multi-body simulation and subsequently optimized using experimental data, as shown in Fig. \ref{fig:pacejka}.
Given the specific application, special care was taken in estimating vertical loads. In particular, the banking effect has been included in the calculation. The exact value is derived from the LiDAR map, subsequently smoothed, and given to the model as a lookup-table function of the position in the track.
Moreover, the model allows to manage the strongly asymmetric setup -- such as camber, toe, inflation pressure, and tire temperature, typically lower on the left side for oval racing -- by defining two different sets of macro-parameters, which differentiate the behavior between left and right turns. This capability can significantly impact multi-vehicle competitions, where overtaking maneuvers can be executed on both the inside and outside of the opponents.
The model is described by the state vector $X$ - Eq. \ref{eq:state}, while the input variables are included in vector $u$ - Eq. \ref{eq:input} 
% Equations for Derivatives of State Variables
The derivative of the state $\dot{X}$ is given by:
\begin{align}
\dot{v}_y &= -v_x \cdot r + \frac{1}{m} \cdot (F_{yf} \cdot \cos(\delta) + F_{yr}) \label{eq:vy_dot} \\
\dot{r}  &= \frac{1}{J_z} \cdot (F_{yf} \cdot \cos(\delta) \cdot l_f - F_{yr} \cdot l_r) \label{eq:r_dot} \\
\dot{a}_y &= \frac{\text{d}}{\text{dt}}\left( \frac{F_{yf} \cdot \cos(\delta) + F_{yr}}{m} \right) \label{eq:ay_dot}
\end{align}
where $m$ and $J_z$ represent the mass and inertia of the vehicle, and $l_f$ and $l_r$ represent the front and rear wheelbase. $F_{yf}$ and $F_{yr}$ are, respectively, the lateral forces generated by the front and rear axles.
% Front Left and Right Wheel Slip Angles
The first step is to compute the slip angles via the congruence equations:

\begin{equation} \label{eq:alpha}
\begin{aligned}
\alpha_{f} &= -\tan^{-1}\left(\frac{v_y + r \cdot l_f}{v_x}\right) + \delta \\
\alpha_{r}    &= \tan^{-1}\left(-\frac{v_y - r \cdot l_r}{v_x}\right) \\
\end{aligned}
\end{equation}

\begin{figure}
	\centering
	\includegraphics[clip, trim=4cm 8.5cm 4cm 9cm, width=1.0\columnwidth]{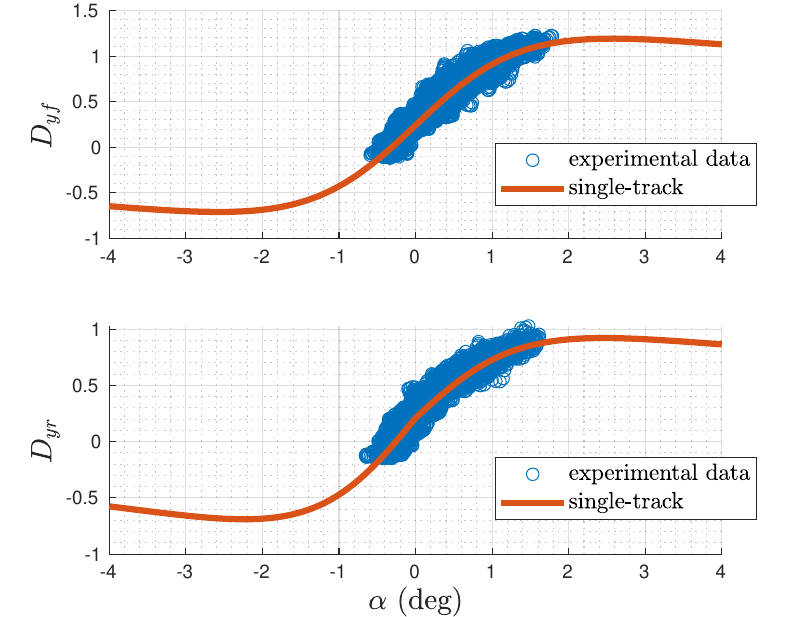}
    \caption{The Pacejka Tire Model (orange line) optimized based on the experimental data (blue scatter) from Texas Motor Speedway, during the event of the Indy Autonomous Challenge.}
	\label{fig:pacejka}
\end{figure}

% Dynamic Lateral Force at Front
Thus, the normalized lateral forces are computed using the slip angles in \ref{eq:alpha}:
\begin{equation} \label{eq:Dy_i}
\begin{split}
D^j_{yi}     &= S_{vyi} + \mu^j_{yi} \cdot \sin\left(C^j_{yi} \cdot \tan\left(B^j_{yi} \cdot \alpha_{i} \right.\right. \\
&\left.\left. - E^j_{yi} \cdot \left(B^j_{yi} \cdot \alpha_{i}) - \tan\left(B^j_{yi} \cdot \alpha_{i}\right)\right)\right)\right) \\
\end{split}
\end{equation}
where the index $i^{th}$ indicates the axle (front, rear), and the index $j^{th}$ indicates the direction of the turn (left turn, right turn). $\mu, B, C, E$ and $S_v$ are the macro-parameters of the Pacejka Magic Formula. In particular, $S_v$ represents a shift parameter needed to take into account the camber thrust that arises due to the setup asymmetry.

% Selection of Dynamic Lateral Force for Front
Hence, two values per axle are computed. The appropriate normalized force is selected depending on the sign of the slip angle, which is considered positive for counterclockwise rotation.
\begin{align}
D^j_{yi}        &= \begin{cases} 
D^l_{yi} & \text{if } \alpha_{i} \geq 0 \\
D^r_{yi} & \text{if } \alpha_{i} < 0
\end{cases} \label{eq:Dy_f}
\end{align}
The normal forces acting on the front and the rear axles are determined by:
\begin{equation} \label{eq:Fz}
\begin{split}
F_{zi}        &= F^{0}_{zi} + F^{aero}_{zi} + F^{b}_{zi} \pm \Delta F^{long}_{z} \\
\end{split}
\end{equation}
where it takes into account gravitational force $F^{0}_{zi}$, aerodynamic effects $F^{aero}_{zi}$, banking contribution $F^{b}_{zi}$, and longitudinal load transfer $\Delta F^{long}_{z}$. The distribution of the normal load between the axles is determined by the positions of the center of gravity and the aerodynamic center of pressure \cite{guiggiani}.
% Banking Force
The normal force due to the banking, influenced by lateral acceleration and $\theta$ angle, is calculated by the equilibrium of the rigid body:
\begin{align}
F^{b}_{z}     &= m \cdot a_y \cdot \tan(\theta) \label{eq:Fz_bank}
\end{align}
% Lateral Forces on Front and Rear
The lateral forces on the front and rear axles, which are critical for vehicle dynamics and handling, are then derived by scaling Eq. \ref{eq:Dy_i} with Eq. \ref{eq:Fz}:
\begin{equation} \label{eq:Fy}
\begin{split}
F_{yf}        &= D_{yf} \cdot F_{zf} \\
F_{yr}        &= D_{yr} \cdot F_{zr}
\end{split}
\end{equation}

Finally, the state space model is used to represent the prediction function 
$f(\mathcal{X}_{k}^{(i)}, u_k)$
already mentioned in \ref{model_update} using an Explicit Euler integrator.

\section{LIDAR SLAM}
\label{section:LiDAR}
The process of LiDAR vehicle localization in autonomous driving incorporates several crucial steps, each contributing significantly to the overall accuracy and reliability of the system. The method used in this paper is an updated version of the one presented in \cite{er-autopilot}.\\
Initially, an offline map is created by methodically driving the vehicle over the track, allowing the collection of detailed LiDAR data. This map serves as a high-quality reference for the vehicle’s localization during autonomous navigation.
The LiDAR Odometry and Mapping (LOAM) technique aligns LiDAR data and GPS data to construct an accurate environmental map \cite{shan2018lego}, \cite{zhang2014loam}. The map undergoes global optimization with frameworks such as GTSAM, enhancing the map's geometry and alignment with GPS trajectories \cite{cadena2016past}. For instance, Figure \ref{fig:slam_map} illustrates the finalized map of the Monza circuit that was achieved.
Online, data from the multiple onboard LiDARs is collected and synchronized, merging the resulting point clouds to form a unified point cloud \cite{xu2021fastlio}. 
Subsequently, the data undergo motion compensation using the measurements from the IMUs, ensuring that vehicular motion does not degrade the quality of the LiDAR data. This compensation is essential to maintain the accuracy of the localization process.
The localization process is characterized by its high degree of accuracy, not only in determining the vehicle's position and orientation but also in estimating velocities. This precision is crucial in dynamic environments where AVs must react swiftly to changing conditions \cite{levinson2011towards}. 
To enhance the robustness of the localization module, the LiDAR data, along with the vehicle's estimated state, is integrated into the EKF mentioned in \ref{lop-ukf-implementation}. The latter, utilizing both LiDAR and GPS data, ensures a more reliable localization by combining multiple sources of information. This integration demonstrates the dual utility of LiDAR odometry for both localization and state estimation. While localization itself is a form of state estimation, the purpose of the UKF in this context is more directed towards control, planning, and understanding vehicle dynamics. It is worth mentioning that no map is required for this approach and only pure odometry could be used; the advantage of the proposed SLAM is high accuracy, comparable to that of RTK-corrected GPS signals.
\begin{figure}
    \centering
    \includegraphics[width=0.675\linewidth]{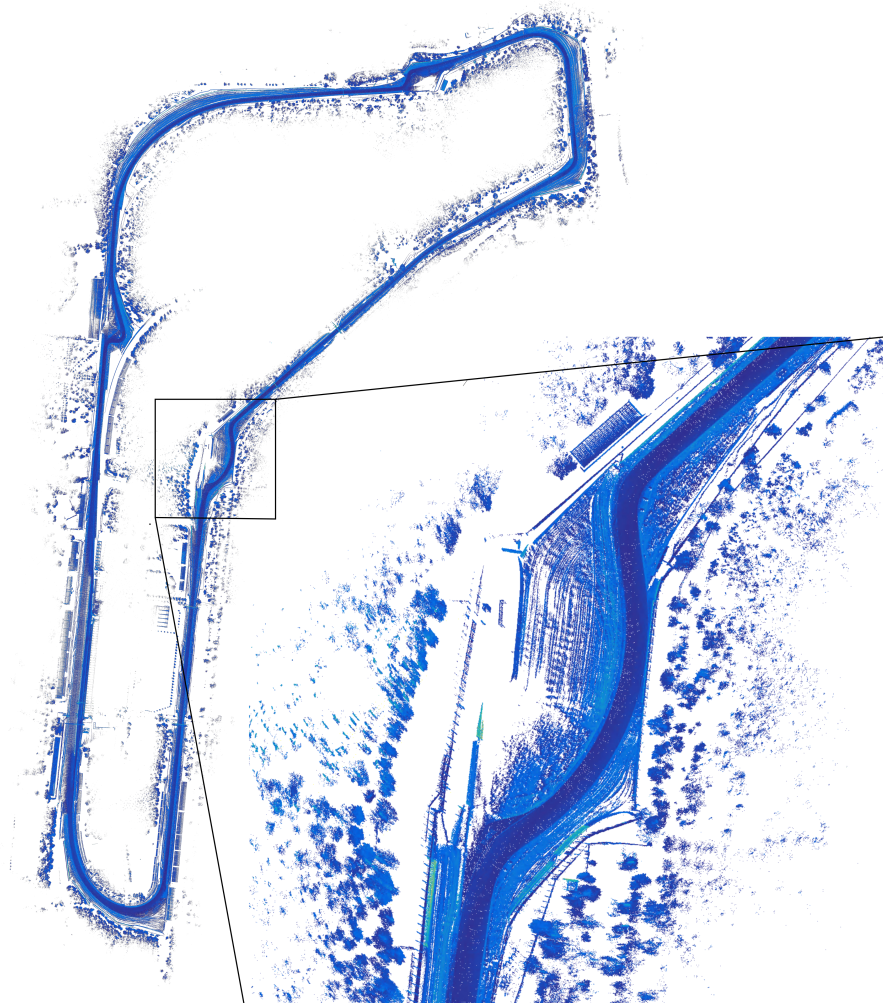}
    \caption{Geo-referenced LiDAR map of Monza F1 track. The zoomed portion shows Variante Ascari more in details}
    \label{fig:slam_map}
\end{figure}

\section{EXPERIMENTAL RESULTS}
\label{section:results}
The results obtained at Texas Motor Speedway (TMS) and at the Monza F1 track configuration during the IAC events are shown in this section, where the LOP-UKF has proven to be robust in both road course and oval speedway scenarios.
The ground truth, provided by the optical sensor, has been filtered with a first-order Butterworth low-pass filter with a cutoff frequency of 5 Hz.
There are currently no online datasets available that contain meaningful data for nonlinear conditions and also include Lidar data. As a result, comparing the results of this approach with other methods on common datasets loses significance. Therefore, the results of this approach will be more anecdotal. However, the main contribution of the proposed method is intended to be its robustness of the solution rather than its mere accuracy.

\subsection{Oval Speedway}
 
\begin{figure}
	\centering
	\includegraphics[clip, trim=4cm 8.5cm 4cm 9cm, width=1.0\columnwidth]{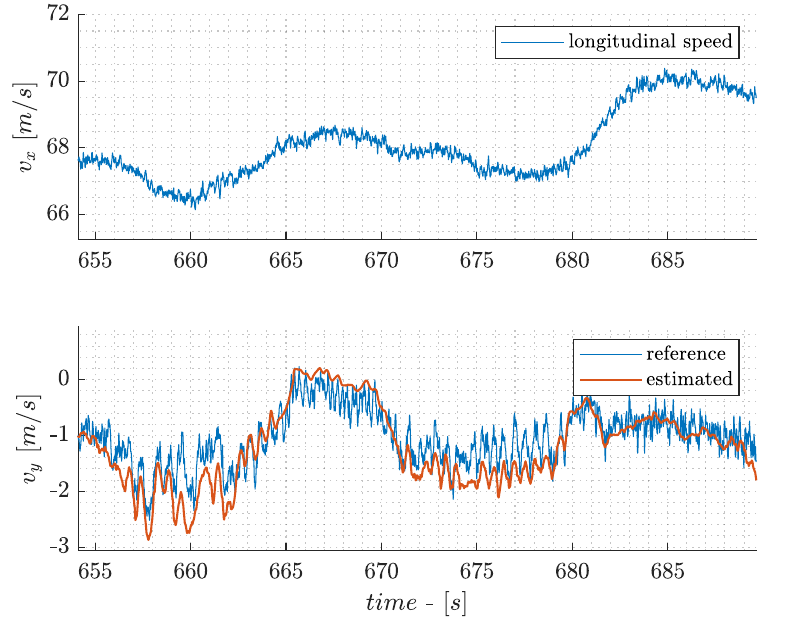}
    \caption{The velocity field of a hot lap in the Oval track of Texas Motor Speedway. The optical sensor serves as a reference.}
	\label{fig:tms_V}
\end{figure}
TMS consists of 4 turns with a radius of about $200 m$. The peculiarity of this track compared to classic ovals is that the banking is variable: in fact, in $turn$ $1-2$ it is around $15^\circ$ while in $turn$ $3-4$ it is approximately $22^\circ$.  Therefore, it is a perfect scenario to validate the power of the proposed solution at high longitudinal speed and different lateral speeds. Indeed, as evident in Fig. \ref{fig:tms_V}, for the same longitudinal speed $\sim 240 km/h$, the lateral speed in curves 1-2 is $\sim 25\%$ higher than in curve 3-4 ($t \sim 675$), reaching a peak of $9 km/h$ ($t \sim 660$).
Despite this, the estimate, as well as the reference, presents a low-frequency oscillation probably caused by the overall dynamics of the system (i.e. steering torque feedback, actuator dynamics, and controller reactivity).
\begin{figure}
	\centering
	\includegraphics[clip, trim=4.5cm 8.5cm 3cm 9cm, width=1.1\columnwidth]{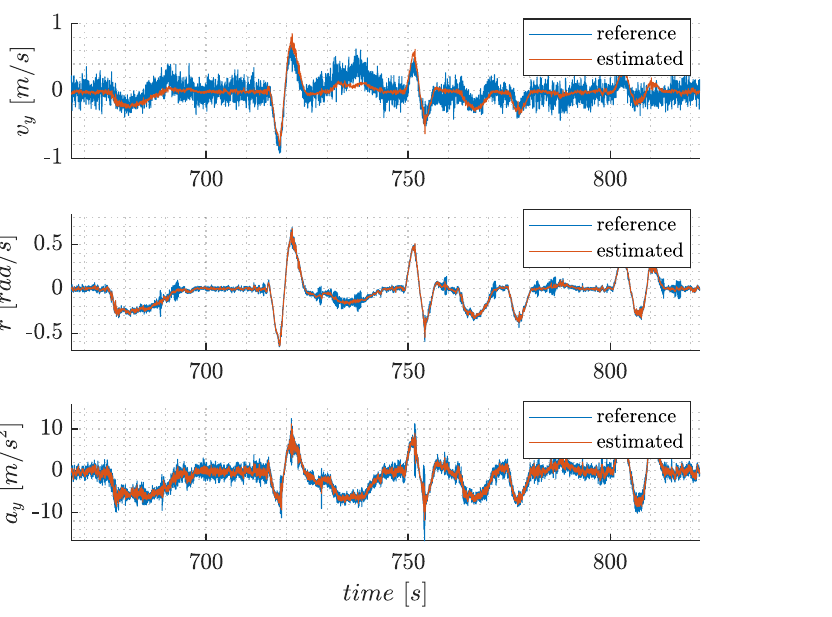}
    \caption{The LOP-filter output compared with the ground truth, provided by the optical sensor and the IMU. The dataset includes a full lap of the Monza track, F1 configuration, starting from Parabolica turn.}
	\label{fig:filter_output}
\end{figure}
\begin{figure}
	\centering
	\includegraphics[clip, trim=4cm 8.5cm 4cm 9cm, width=1.0\columnwidth]{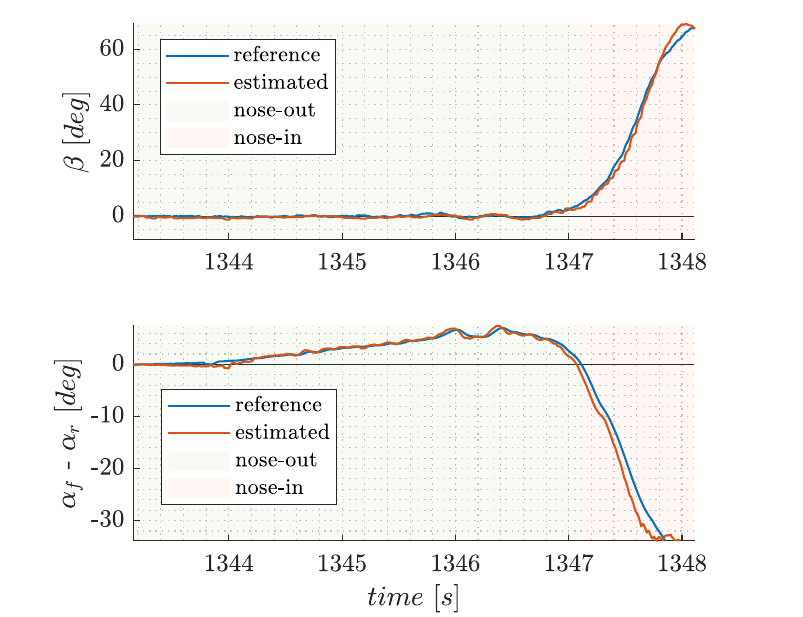}
    \caption{In the upper plot, the sideslip angle is shown. In the lower plot, we present the understeer degree, a very usefull indicator of the car behavior in the curvilinear motion (positive values are associated to nose-out).}
	\label{fig:beta_vs_k_crash}
\end{figure}
\begin{figure}
	\centering
	\includegraphics[clip, trim=4.5cm 11cm 2cm 11cm, width=1.2\columnwidth]{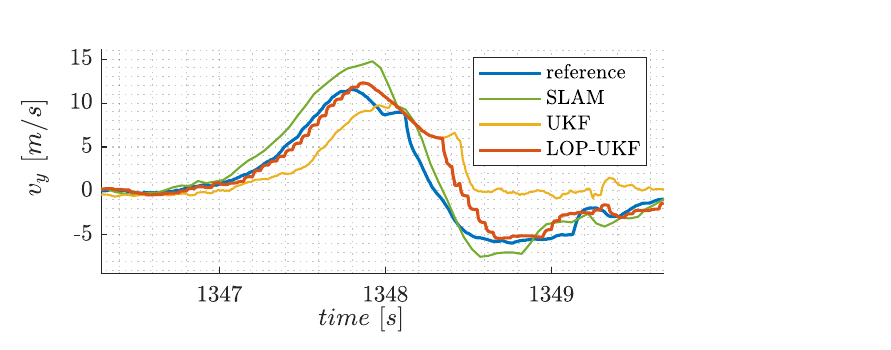}
    \caption{The figure shows the filter output with only the model (ochre line) and combined with the LiDAR measurement (orange line). Kistler sensor serves as reference.}
	\label{fig:slam_cov}
\end{figure}
It is also important to note that in the linear zone of the tire, the trend and value during transients are highly accurate. However, it becomes evident that during cornering, where tires enter the nonlinear zone, the numerical value of $v_y$ deviates slightly. This is mainly due to the uncertainty of the parameters in the Pacejka model, which were optimized following this test, as demonstrated in Fig. \ref{fig:pacejka}. Indeed, in this experiment the filter is running with the first attempt parameters derived by the simulator.

\subsection{Road Course}
In Fig. \ref{fig:filter_output} a full lap of the Monza circuit is shown. Beside the performance not close to the tire limit, all three outputs of the LOP-UKF match well with the ground truth. It is important to highlight that with such a simplified model, the filter is incapable of effectively capturing both tight turns and wide turns. With the chosen parametrization, it produces an extremely accurate estimation in tight turns - e.g., Prima Variante, $t \sim 720$ - but it is penalized in wide turns - e.g., Biassono turn, $t \sim 735$.
Fig. \ref{fig:beta_vs_k_crash} and Fig. \ref{fig:slam_cov}, depict the results under wet asphalt in the Parabolica turn of Monza, thus the model parameters defined in \ref{section:pacejka}, under dry track, lacks of accuracy. This is a perfect scenario for demonstrating the efficacy of our method under extreme conditions. The estimation of $v_y$ and $r$ enables us to calculate the understeer degree, which provides an essential indicator of the handling and safety of the car. As visible in \ref{fig:beta_vs_k_crash}, in the turn-in phase, the car exhibits a significant nose-out \cite{bergman} (green background), which turns into a nose-in (orange background) until the car spins. While the UKF relying just on the model reproduces the trend but with a significant absolute error, the LOP-UKF is able to capture the car dynamic with an overall high accuracy.
It is important to note that the final estimate depends on a trade-off between the confidence that is given to the measurements and the model. This must be properly defined to avoid replicating the SLAM peaks, visible in figure \ref{fig:vy_compare}, which are caused by errors on the map and platform vibrations. However, it is essential to take into account the correctness of the model parameters, which can vary significantly depending on the evolution of the car or the track, as observed in the aforementioned experiment.

\section{CONCLUSION}
The experimental results demonstrate that integrating LiDAR odometry with an UKF, based on the Pacejka tire model, offers a robust method for estimating vehicle lateral velocity. The effectiveness of LOP-UKF was proven across various tracks and grip conditions. Previous discussions in Section \ref{section:related} highlighted that the literature contains numerous valid approaches for estimating sideslip in different contexts. The accuracy of state-of-the-art solutions is largely dependent on the precision of the utilized model, which, in turn, relies heavily on empirical data for calibration. This principle applies universally to data-driven methods, such as NN and mathematical tire models. Our LOP-UKF can provide precise estimations even with low confidence in the LiDAR input, contingent upon the tire parameters being calibrated under grip conditions similar to those tested. Although purely mathematical models offer the flexibility to adjust parameters for various vehicle setups or entirely different vehicles, such adaptations require experimental data. Consequently, the outcomes of the SLAM process significantly bolster the effectiveness of both EKF and LOP-UKF, especially when vehicle dynamics deviate substantially from the models employed in these filters. These insights suggest that future research on sideslip estimation could benefit from exploring other hybrid approaches to enhance robustness. Real-time adaptable tire models may provide a solution for generalizing estimators, but true generalization necessitates diverse sensor inputs that offer independent measures or estimates, regardless of the filter's model. The use of cameras and radars has become commonplace in many vehicles. Additionally, with the advent of more affordable LiDAR technology, previously deemed expensive, certain car manufacturers began its integration into production vehicles. This trend highlights the potential for new hybrid approaches in state estimation, aiming to enhance the performance of ADAS.
\begin{figure}
	\centering
	\includegraphics[clip, trim=4.5cm 11.3cm 2cm 11.5cm, width=1.2\columnwidth]{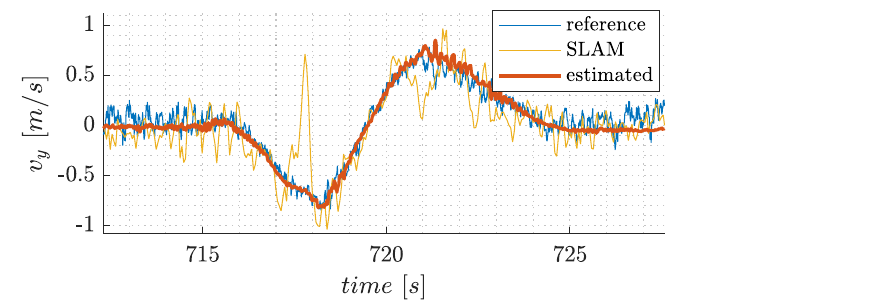}
    \caption{Lateral velocity comparison: the reference signal is provided by the optical sensor, the slam measure is provided by the LiDAR, and the estimated is the LOP-filter output.}
	\label{fig:vy_compare}
\end{figure}

%%%%%%%%%%%%%%%%%%%%%%%%%%%%%%%%%%%%%%%%%%%%%%%%%%%%%%%%%%%%%%%%%%%%%%%%%%%%%%%%

\end{document}